# Image Fusion in Remote Sensing: An Overview and Meta Analysis

Hessah Albanwan, Rongjun Qin, Yang Tang

*Abstract*— **Image fusion in Remote Sensing (RS) has been a consistent demand due to its ability to turn raw images of different resolutions, sources, and modalities into accurate, complete, and spatio-temporally coherent images. It greatly facilitates downstream applications such as pan-sharpening, change detection, land-cover classification, etc. Yet, image fusion solutions are highly disparate to various remote sensing problems and thus are often narrowly defined in existing reviews as topical applications, such as pan-sharpening, and spatial-temporal image fusion. Considering that image fusion can be theoretically applied to any gridded data through pixel-level operations, in this paper, we expanded its scope by comprehensively surveying relevant works with a simple taxonomy: 1) many-to-one image fusion; 2) many-to-many image fusion. This simple taxonomy defines image fusion as a mapping problem that turns either a single or a set of images into another single or set of images, depending on the desired coherence, e.g., spectral, spatial/resolution coherence, etc. We show that this simple taxonomy, despite the significant modality difference it covers, can be presented by a conceptually easy framework. In addition, we provide a meta-analysis to review the major papers studying the various types of image fusion and their applications over the years (from the 1980s to date), covering 5,926 peer-reviewed papers. Finally, we discuss the main benefits and emerging challenges to provide open research directions and potential future works.**

## 1. INTRODUCTION

IMAGE fusion can be broadly defined as the process of combining information from single or multiple images into more accurate, complete, and spatiotemporally coherent images. It has received great attention in remote sensing (RS) over the years due to the ever-growing development of sensors and platforms, which provide a nearly unlimited archive of earth observation data. Complementary images from different sensors/platforms enable augmented observations of RS data to enrich and expand the knowledge around them [1], [2], [3], [4]. Existing image fusion methods that fundamentally achieve augmented observations were shown to be of great interest in many RS applications [1], [2], [3], [4], [5], [6], [7], [8]. These methods aim at enhancing spatial, temporal, and spectral resolutions at the image level and, at the same time, improving the learnable features from various sources to achieve better image interpretation performances [1], [2], [4]. For instance, by combining low-spatial high-temporal resolution images from MODIS with high-spatial low-temporal resolution images from Landsat, we can generate high-spatial and high-temporal images for more accurate land-cover and biomass monitoring applications [5], [6], [7], [9]. In contrast, high-spectral low-spatial resolution data (e.g., multi- / hyper-spectral images) can be used to enhance the spectral resolution of panchromatic images, i.e., pan-sharpening [6], [8], [10]. Beyond single-modal data (i.e., optical images), researchers have also explored using multi-modality data for image-based feature learning [11], [12]. These multi-modality data include but are not limited to optical images, LiDAR (Light Detection and Ranging), and SAR (Synthetic Aperture Radar) data, which have shown that such a feature-level fusion can significantly enhance the performance of machine learning tasks for remote sensing [13], [14], [15].

The past literature has provided syntheses of the existing image fusion efforts with a great focus on individual topical applications and their specific fusion algorithms, for example, pan-sharpening, super-resolution, or feature-level fusion for classification [1], [16], [17], [18]. Their limited area of investigation restricts comprehensive understanding and views of the image fusion concepts, which can also be applicable to other RS tasks and problems. For instance, with some adaptation, fusion techniques used in pan-sharpening can be applied to super-resolution or spatiotemporal inference problems. In RS, the majority of data is still operating through a raster format defined under a geo-referenced coordinate frame (i.e., projection space) [19], [20]. On the one hand, this made it relatively easy to register data with respect to each other and, on the other hand, made it highly effective to implement algorithms to two-dimensional (2D) grids where multiple data layers can be fused. Hence, fusion applications on 2D image grids can be easily extended through various RS applications and data products, such as optical images, classification maps, thematic maps, or Digital surface models (DSMs) [1], [2], [3], [4], [5], [6], [7], [8]. These products, although being presented as 2D raster, possess completely different content associated with different uncertainties and numerical values, in addition to being used in completely different applications. Therefore, it can be beneficial to comprehensively survey image fusion methods that are

The study is partially supported by the Office of Naval Research (Grand No. N000142012141 & N000142312670). (Corresponding author: Rongjun Qin, qin.324@osu.edu).

Hessah Albanwan is with the Geospatial Data Analytics Lab and Department of Civil, Environment and Geodetic Engineering, The Ohio State University, Columbus, USA (email: albanwan.1@osu.edu)

Rongjun Qin is with the Geospatial Data Analytics Lab, Department of Civil, Environment and Geodetic Engineering and the Department of Electrical and Computer Engineering, and Translational Data Analytics Institute, The Ohio state University, Columbus, USA (email: qin.324@osu.edu)

Yang Tang is with the Geospatial Data Analytics Lab and Department of Civil, Environment and Geodetic Engineering, The Ohio State University, Columbus, USA (email: tang.1693@osu.edu)

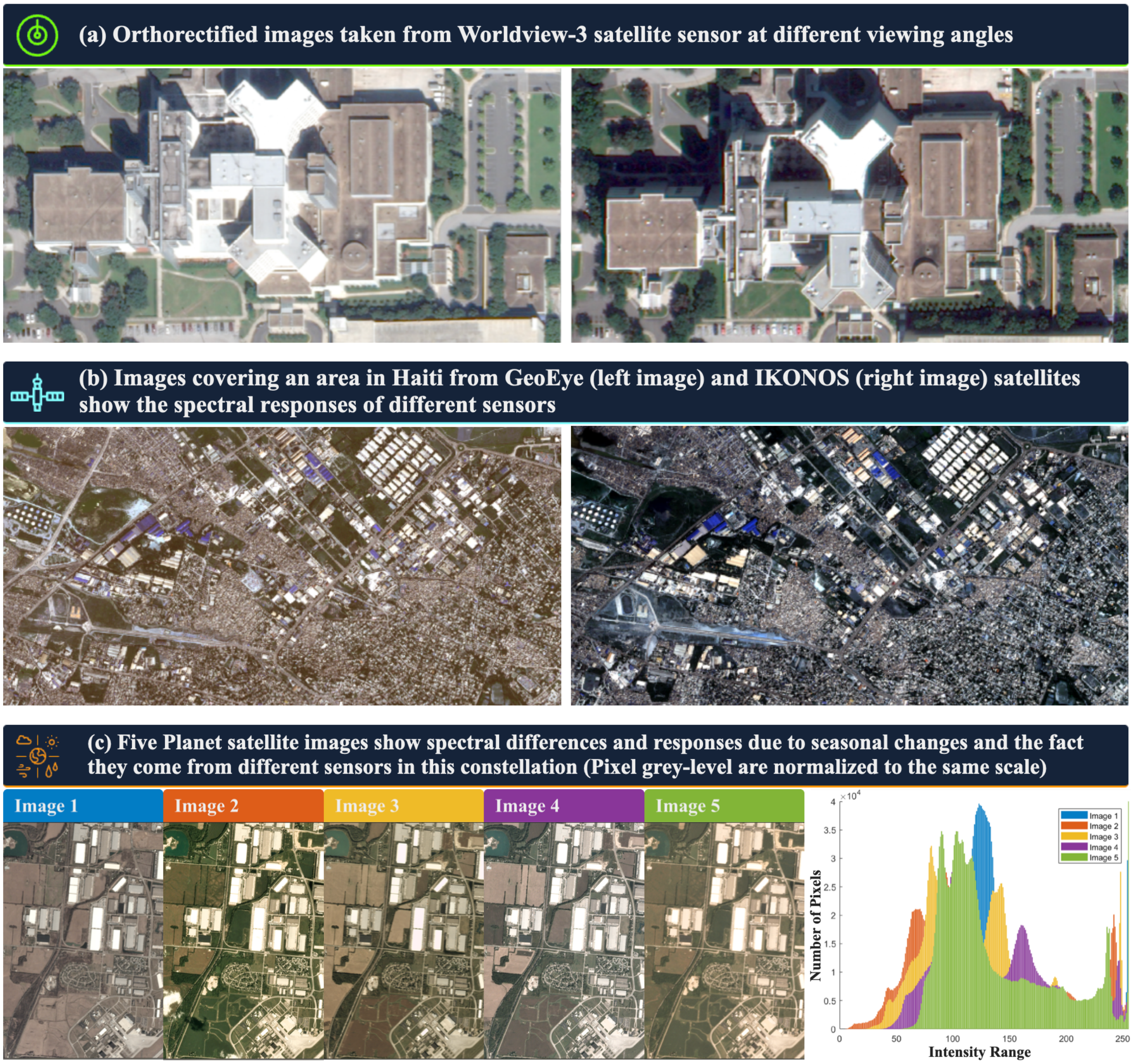

**Figure 1.** Examples of radiometric inconsistencies between RS images due to differences in (a) viewing direction, (b) sensors, and (c) season.

generally applicable to various applications with a broader scope.

This systematic review focuses on providing a comprehensive revision and a meta-analysis on the topic of image fusion in the RS field. Despite the significant difference in the type of image fusion and its applications, we consider that their fusion methods can share similar concepts. Therefore, the framework used in one method can be used to augment other types of images and applications. For example, fusion techniques used for super-resolution can also be applied for pan-sharpening, while methods used to process and fuse time series analysis, such as filtering, can be used for image pre-processing. Moreover, publications informed by the meta-analysis are used as measuring entities to evaluate current scientific activities such as fusion algorithms and applications. We hereby conduct a search in Scopus for all original peer-reviewed publications on the subject of image fusion in RS published between 1980 and 2023. We reviewed 5,926 articles related to image fusion and its applications in RS, covering specific fusion algorithms and downstream applications in studying urban and environmental issues. Therefore, we expand the scope of image fusion, as practiced in the RS community, by providing a simple yet comprehensive taxonomy that defines fusion as a mapping function operating on different numbers of images to output single or multiple images. We also show that despite the significant "input-output" difference, the fusion problem can be explained by a conceptually easy framework. The meta-analysis plays a significant role in providing insights into the most recent fusion techniques and current applications, specifically the learning-based and deep learning methods, which have become very popular and require a literature update.

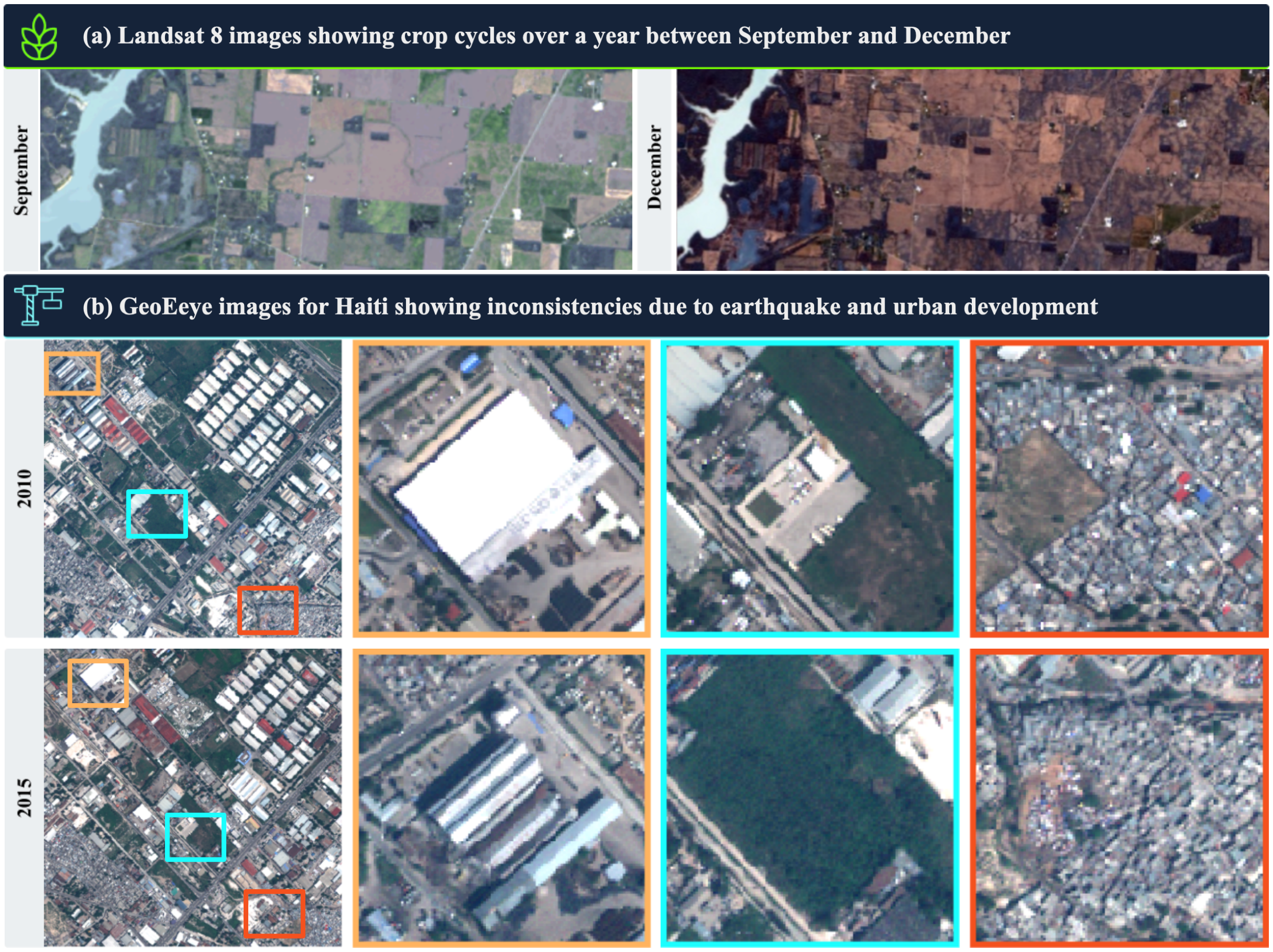

**Figure 2.** Examples of spatiotemporal inconsistencies are shown in RS images due to (a) natural causes, such as crop cycles, and (b) natural and human causes, like earthquakes and new constructions.

The rest of the paper is divided into the following Sections: Section 2 introduces the background information of RS image fusion, which includes the main challenges of image fusion in the RS field, the proposed taxonomy of image fusion, and the commonly used image fusion approaches. Section 3 describes the method and guidelines for this meta-analysis to review image fusion and its evolution in the RS field over the years. Section 4 presents the meta-analysis results and statistical information concerning the number of publications, applications, methods, etc. Finally, Section 5 summarizes this review with an open discussion on the main challenges and potential future works of RS image fusion.

## 2. An Overview of Remote Sensing Image Fusion

Despite current RS technologies and sensor advances, image fusion still faces many challenges. Factors of concern such as the image quality (e.g., noisy level, blurriness, radiometric imbalances), the complexity of the scene (e.g., convoluted dense urban and greenspace, homogeneous vegetated regions with high seasonal differences), and processing algorithms (e.g., modeling errors and propagated errors) play important roles fusion solutions. Therefore, understanding these contributing factors is essential to comprehend the fusion problem and the capabilities of its algorithms. In this section, we discuss the main factors impacting image fusion quality, the proposed taxonomy that categorizes fusion into two groups based on current practices, and finally, the fusion techniques currently in use for various RS applications.

### 2.1. Factors Impacting the Quality of Remote Sensing Image Fusion

As mentioned in Section 1, a typical RS image is a 2D raster data representing physical information about a surface area on earth. The raster "data" can be of any type, such as optical images, elevation maps (e.g., Digital Elevation Model (DEM), Digital Surface Model (DSM), and Digital Terrain Model (DTM)), classification maps, thematic maps, radar images, etc. The pixel information of these images can be directly obtained from sensors (e.g., radiometry) or derived from specific algorithms (e.g., elevation maps from photogrammetry methods). In both cases, inconsistency can be developed due to the varying conditions (e.g., acquisition angle, weather, and season) or errors propagating through the processing algorithms (e.g., DSM errors), leading to low-quality input prior to fusion. This section summarizes three crucial factors, including radiometric inconsistency, spatiotemporal consistency, and derived/propagated errors from image products such as elevation maps. Details are in the following sections.

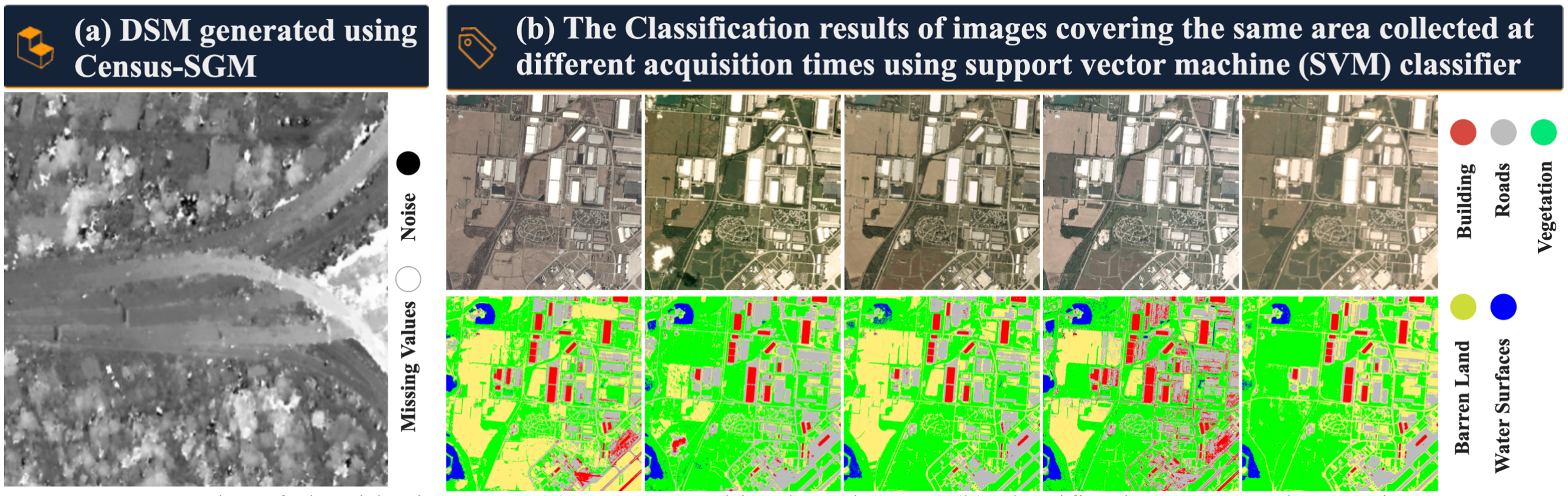

**Figure 3.** Examples of algorithmic errors: (a) stereo matching based DSM; (b) classification maps under varying illumination conditions.

### 2.1.1. Radiometric Inconsistency

Radiometric inconsistency refers to heterogeneous spectral information of objects when observed by different images under varying acquisition conditions [21]. For instance, Figure 1(a) shows the same objects observed by images under different sun angles and sensor's viewing directions, where corresponding pixels show different brightness and chromaticity. The sensors themselves, due to different calibration and spectral responses, may also produce spectral heterogeneities [22], [23], [24]. As an example, Figure 1(b) shows two images for an area in Haiti taken by GeoEye (original Ground Sampling Distance (GSD): 0.5m) and IKONOS (original GSD: 1m), resampled to the same GSD (1m), which still show very different contrast and saturation. Phenological changes, such as those affected by atmospheric conditions and seasons, may also produce spectrally variant images [25], [26]. Figure 1(c) shows an example of five Planet satellite images capturing the same scene within one year of time, where there are significant appearance differences among different collections. Because most fusion algorithms expect spectrally homogenous images, this problem may lead to incorrect use of pixel information data, leading to low-quality fusion results.

### 2.1.2. Spatiotemporal Inconsistency

The spatiotemporal inconsistency of RS data refers to images covering the same geographical area but presenting different scene content due to land cover changes over time [27]. This type of inconsistency can sometimes be convoluted by the radiometric inconsistency caused by sensors (explained in 2.1.1.) since images capturing a scene of a different time may come from a different sensor in one or multiple satellite constellations. Typically, spatiotemporal inconsistencies are considered due to natural causes such as seasons, phenological cycles, disasters, etc., or anthropogenic activities such as urban developments, wars, demolitions, etc. An example of the phenological growth of crops is shown in Figure 2(a), in which we can see the vegetated area in the suburban area in September turned into barren land in December, leading to significant appearance changes. Figure 2(b) shows another example of the spatiotemporal inconsistency, depicting the differences caused by the earthquake (Haiti earthquake in 2010) and its status in 2015. It is obvious that there have been new buildings established to accommodate refugees in houses or shelters (See Figure 2(b)-red box), as well as new buildings (See Figure 2(b) blue and yellow boxes). These inconsistencies, on the one hand, tell the surface dynamics, while on the other hand, when these images are directly used for super-resolution, it will undoubtably produce haunted effects. Therefore, strategies for dealing with or avoiding these inconsistencies are necessary for successful fusion applications.

### 2.1.3. Propagated Error

Processed RS data, e.g., classification maps or stereo-based DSM, may contain errors during the data processing [28], [29]. For example, a typical DSM generation pipeline includes steps such as epipolar rectification, stereo matching, and triangulation from point clouds, which may accumulate errors due to numerical and modeling approximation [30], [31] in these steps. An example DSM generated using a semi-global matching (SGM) algorithm [32] is shown in Figure 3(a), where visible noises can be observed on both flat surfaces and building boundaries. Such errors are typical and can be algorithm-dependent, for example, certain stereo-matching algorithms produce sharper boundaries but produce salt-and-pepper noises (such as using normalized cross-correlation algorithms [33], while algorithms that enforce smoothness (e.g., global matching algorithms, such as SGM and deep learning methods (e.g., PSMNet [34])) will produce less salt-and-pepper noise, but may omit important and sharpen structures, such as small and tall buildings. Therefore, understanding such characteristics as priors for fusion is beneficial. Following a similar consideration, when classification maps are used for fusion, errors from classification maps may yield inconsistent fusion results. For example, Figure 3(b) shows classification maps predicted from images captured at different times with only days of difference, where the illumination differences result in different noise levels, sometimes quite substantial (i.e., a large portion of the Barren land is mixed with the Vegetation class over the short time). The errors introduced by these learning models are still not very well characterized, which can also bring significant challenges for fusion algorithms.

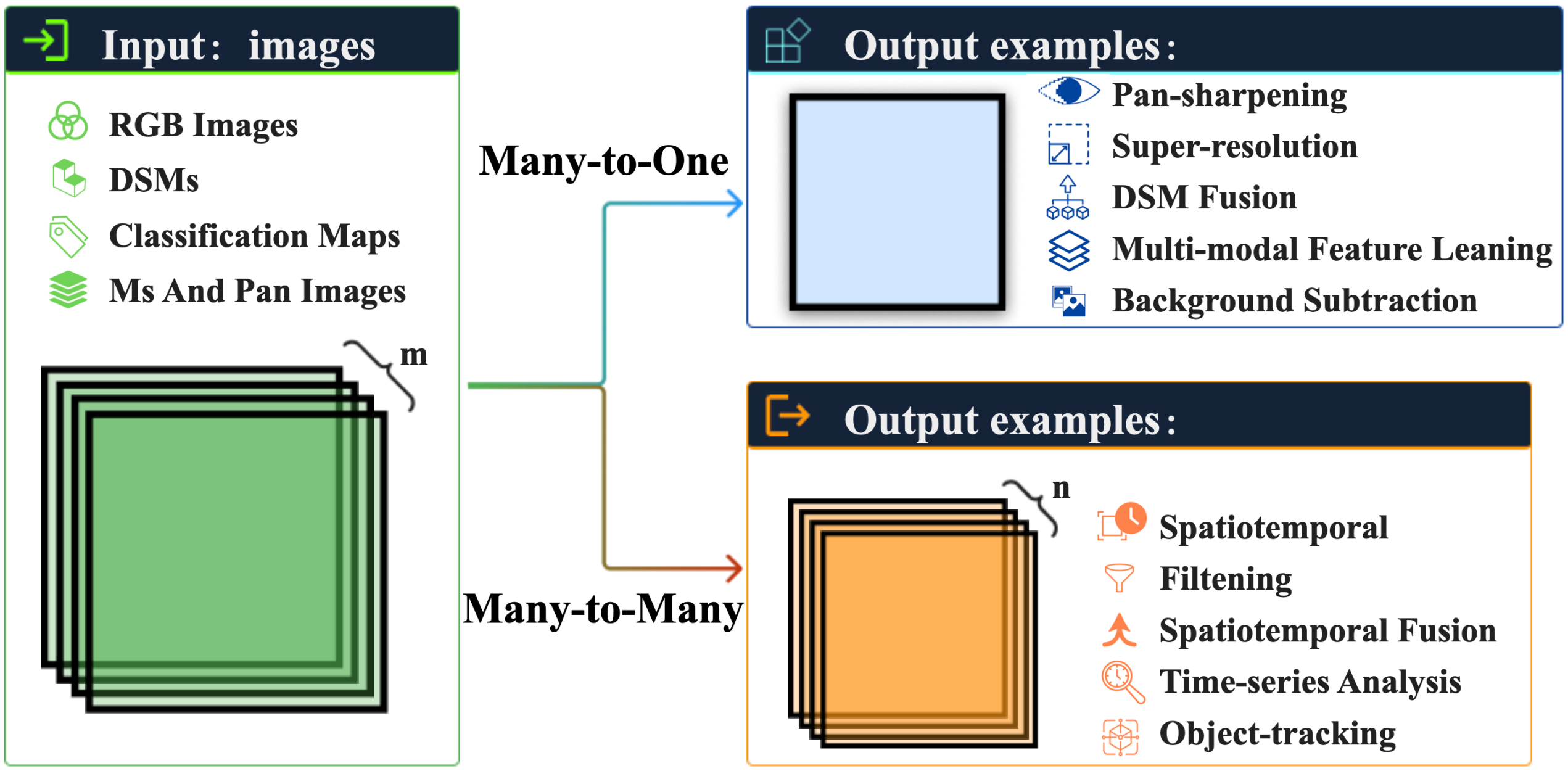

**Figure 4.** An overview of the types of image fusion in remote sensing. Note: 'm' and 'n' in the figure denote the numbers of input and output images, respectively.

## 2.2. A Generic Taxonomy of Remote Sensing Image Fusion

Image fusion is a highly disparate topic that can be applied to various RS tasks. In the current literature, these different fusion tasks are often narrowly defined within their own methodological domain. For example, pan-sharpening is often considered an independent task other than super-resolution. However, many of these fusion tasks are of a similar nature, and their methodologies share common strategies. The generic fusion problem can be described mathematically: given a set of observations X={x_1,x_2,x_3,⋯,x_n ┤ | x_i∈R^(w,h,c_x )}, where w,h, and c_x refer to the width, length, and number of channels for raster data. Thus, X can be represented as a tensor belonging to R^(w×h×c_x×n) . A simple linear fusion algorithm aims to convolute the tensor X , through each individual observation via a mapping function f(X) , and output another tensor Y={y_1,y_2,y_3,⋯ ,y_m ┤ | y_i ∈ R^(w,h,c_y )}, where c_y refers to the number of channels for each one of out of the m outputs, thus Y can be represented as a tensor belonging to R^(w×h×c_y×m) as follows:

$$\boldsymbol{Y} = \boldsymbol{f}(\boldsymbol{X}) \quad (1)$$

This is easily extendable when f(·) is a non-linear function involving more complex and multi-step mappings [4], [16], [18]. Here, depending on m = 1 or more, we categorize these strategies into two with the simplest "input-output" fashion, being 1) "many-to-one", meaning that several input images are fused into a single image output representing enhanced results, where these several inputs often refer to multiple homogenous observations and the fusion of which refers to output with enhanced spatial/spectral resolution or accuracy. 2) "many-to-many", meaning that several input images are fused into several (often equivalent number of) outputs with enhanced spatiotemporal resolution. Details of these two types will be further explained in the following subsections. As mentioned earlier, we divided the RS image fusion problem into two types: many-to-one (M2O) and many-to-many (M2M). Figure 4 provides an overview of these types of image fusion.

### Many-to-one (M2O) Image Fusion

Many-to-one image fusion refers to the process of combining several images into a single image. This type of problem serves many applications, such as pan-sharpening, super-resolution, DSM fusion, multi-modal feature learning for classification, background image generation for tracking, etc. [7], [12], [35], [36], [37], [38], [39], [40], [41], [42], [43], [44], [45]. These fusion problems have been intensively investigated over the years [1], [2], [4]. The general premise is that the information of different inputs residing in the same pixel location should refer to the same static object with the same or complementary modality. Therefore, the goal of the fusion can be generally categorized into three aspects: 1) estimate a more accurate observation of the same pixel, using the redundant pixel information from these multiple inputs (e.g., super-resolution, Pan-sharpen, and DSM fusion) [8], [42], [43], [46], [47], [48], [49], [50], [51]; 2) generate a background image to represent the static scene in preparation for object tracking [52], [53], [54], [55], [56], [57]; 3) extract features from different modality (different spectral bands, height, microwave reflectance) to achieve enhanced classification results to generate accurate thematic maps [36], [40], [44], [58], [59], [60], [61], [62], [63].

**Many-to-many (M2M) Image Fusion**

Many-to-many image fusion combines the knowledge/information from multiple images by simultaneously or sequentially processing them to yield a group of refined images, often matching the number of inputs. This type of fusion is mostly used for applications that require a bundle of enhanced images, such as multi-image filtering, change detection over time, Multiview 3D reconstruction, spectral coherence between a group of images, and constant monitoring of the earth's surface. [25], [44], [58], [64], [65], [66], [67], [68]. The general premise of this type of image fusion is that the information of different inputs residing in the same pixel location represents a certain state of the objects in that pixel, which are complementary to each other, such that these inputs can be used to augment others and produce refined outputs that have more accurate pixel information per output. For example, spatiotemporal filtering approaches aim to denoise individual images using inputs at the temporal direction, such that pixels polluted by atmospheric conditions can be recovered using other images acquired under a better atmospheric condition. Thus, similar to the M2O category, it exploits spatial and temporal correlations between images, which are the key ingredients to providing accurate and reliable fused images. Examples of applications in this category can be spatiotemporal fusion to achieve an augmented spatial and temporal resolution of an image sequence, satellite video data augmentation, object tracking, and time-series analysis [53], [57], [69], [70], [71], [72], [73], [74].

### 2.3. General Fusion Approaches in Remote Sensing

Image fusion approaches have been developing rapidly over the years to tackle the fundamental challenges mentioned in Section 2.1 and fuel various RS applications. Methods supporting both M2O and M2M fusion can range from simple arithmetic operations to complex learning-based approaches (i.e., machine learning (ML) and deep learning (DL) methods). The choice of fusion approach mainly depends on the type of integrated images and applications. For example, combining multimodal or multisource images (e.g., RGB image and DSM) to synthesize mixed modality data requires adaptation of differences in image structures, viewing perspectives, resolutions, platforms, etc. Feature-level fusion integrates these diverse modality data into a common feature space prior to fusion. In contrast, unimodal image fusion, due to the modality information, can be directly fused (through linear operations) and often starts from pixel-based operations (e.g., averaging). The RS applications themselves define the necessary level of accuracy for the fusion algorithm. For example, averaging multitemporal images can be sufficient to achieve moderate accuracy for scene-level information extraction (e.g., crop yields prediction) while performing per-pixel image interpretation (e.g., semantic and panoptic segmentation) for urban scenes with buildings and trees of different scales, weighted averaging or learned-fusion at various levels can be more effective [1], [58], [75], [76]. Understanding these factors is essential to developing an efficient fusion algorithm that can adapt to different images, scenes, and applications.

Generally, fusion methods can be categorized broadly into two groups: classic arithmetic methods (classic method hereafter) and deep learning-based methods (learning-based method hereafter). Classic fusion methods perform basic operations to combine images based on statistical, numerical, and domain transform methods [25], [69], [73], [77], [78], [79], [80], [81]. These methods (mostly based on linear operations) are primarily based on pixel-based and are operated locally (e.g., color transformation from RGB to Intensity-Hue-Saturation (IHS)). Classic fusion methods are not data-driven and do not require complex structures to formulate the fusion problem; thus, they are considered simple to implement and efficient to run. However, the simplicity of the classic methods (such as window-based methods) may have limited performance when processing complex features such as thin structures or repeated textures [25], [82]. Moreover, they assume fixed and known data and noise distribution (e.g., normally distributed) and thus may not adapt well to different data types, scene contents, and distributions. On the other hand, the learning-based methods have recently shown remarkable performances in many image fusion tasks [83], [84]. Examples of these methods include classic statistical ML and DL methods [36], [38], [44], [47], [62], [65], [66], [85], [86], [87], which are found to outperform many of the classic fusion methods due to their ability to extract complex features and patterns from images. Like many existing learning-based approaches, adaptivity and generalization are considered a challenge unless relevant, high-quality input training data is available, which can be very expensive considering the geographically diverse domain of RS images. In this section, we will review both types of techniques under the context of image fusion, and we will focus on describing the most common methods and highlighting their unique advantages and limitations. A summary of Image fusion methods is shown in Table I.

#### 2.3.1. Classic Image Fusion Methods

Classic image fusion methods are defined as mostly non-learning approaches, which are based on statistical, numerical, and domain transformation methods. These methods have a simple structure and do not require complex modeling of the fusion problem or learning from large datasets; thus, they are relatively easy to implement and considered computationally fast, and for a long time, they serve as the mainstream methods for practical uses. Most of these methods perform local operations, i.e., informing from neighboring pixels in space or time for noisy removal and purifying the fused results. Figure 5 shows an overview of the classic image fusion, where several DSMs are fused using median filtering and per-pixel operations to update every pixel using information from their neighbors. Depending on how the local pixel information is used, the classic fusion methods are further divided into three categories: 1) Statistical/probability-based methods, which mostly assume linear operations on local pixel values to yield fused results; 2) Prediction-based methods, which use designated functions to explicitly model the mapping between the input pixel values and the fused values. 3) Transformation-based fusion, which fuses information under a transformed space.

**Table I. Image fusion methods in remote sensing**

| Type of Method | M2O | M2M |
|---|---|---|
| **Classic Image Fusion Methods (Section 2.3.1)** | | |
| ***Statistical/probability-based methods*** | | |
| • **Statistical filtering** | | |
| Basic statical methods: | | |
| - Average | [37], [88] | [89] |
| -Median | [90] | N/A |
| - Standard deviation | [91] | [92] |
| - Histogram | [80] | [78] |
| Spatiotemporal filtering: | | |
| -Bilateral filter | [93] | [25] |
| - Spatiotemporal filter | [9] | [58], [69] |
| -Statistical weighting | [43] | [94] |
| -Image-guided filters | [75], [95] | [96], [97] |
| • **Probabilistic modeling** | | |
| Maximum a-posteriori | [98] | [99] |
| Bayesian fusion | [100], [101] | [102] |
| Dempster-Shafer theory | [54] | [103] |
| ***Prediction-based methods*** | | |
| • **Interpolation/ extrapolation** | | |
| Spline | [104] | [105] |
| Bicubic | [106] | |
| Kring | [107] | [108] |
| **Regression** | | |
| Linear regression | [109] | [71], [74], [110] |
| Least squares method | [111] | [74], [112] |
| **Partial differential equations** | [113], [114] | [115] |
| **Optimization** | | |
| Iterative methods | [109], [116] | [58] |
| ***Transformation-based fusion*** | | |
| • **Spatial domain transformation** | | |
| **Color transformation** | | |
| - Intensity-Hue-Saturation | [117] | [118] |
| - Brovey transformation | [119] | [120], [121] |
| **Projection and substitution** | | |
| - Principal Component Substitution | [122] | [123] |
| - Gram-Schmidt | [7] | [124] |
| • **Frequency domain transformation** | | |
| Fourier transform | [125], [126] | [127], [128] |
| Discrete Fourier transform | [129], [130] | [131] |
| Discrete Cosine transform | [132], [133] | [134] |
| • **Wavelet transformation** | | |
| Continuous wavelet transform | [88], [135], [136] | [137], [138] |
| Discrete wavelet transform | [79], [139], [140] | [141] |
| • **Multiresolution analysis (MRA)** | | |
| Curvelet transform | [142] | [143], [144] |
| Laplacian pyramid transform | [5], [145] | N/A |
| **Learning-based Image Fusion Methods (Section 2.3.2)** | | |
| ***Conventional machine learning methods*** | | |
| Support vector machine | [61], [146] | [147] |
| Random forest | [148], [149] | [150] |
| K-means clustering | [151] | [152] |
| ***Deep learning methods*** | | |
| **Deep learning for spatial-spectral image fusion** | | |
| Convolutional neural networks | [153], [154], [155] | [156], [157] |
| Residual Network (ResNet) | [47], [158] | [159] |
| U-Net | [160], [161] | [162] |
| • **Deep learning for spatiotemporal image fusion** | | |
| Recurrent neural networks | [163] | [84] |
| Long short-term memory | [164], [165] | [166] |
| Gated recurrent unit | [86], [167] | [168] |
| Attention mechanism | [87], [169] | [170] |
| Transformers | [171], [172] | [173], [174] |

### Statistical/probability-based Methods

Statistical methods explore the statistical properties of pixel values within a local window (spatially and/or temporally). Examples of these methods include using mean, median, standard deviation, and histograms as means to compute the updated pixel values. For a long time, statistical filter-based methods have dominated the fusion approaches [25], [75], [175], [176] due to their simplicity and the ability to easily scale to large-format remote sensing images. Given the input of a stack of images, statistical filters produce smoothed images and suppress unwanted noise during the fusion process, in which the statistical quantities play the role (i.e., mean, median, etc.). The window size was considered a hyperparameter, which may affect the fusion results depending on the noise level of the input. A tangible strategy is to create "soft" weighting for pixels within a window to reflect the contribution of each pixel in the fusion process. To adaptively determine the weight of each pixel, statistical quantities, such as entropy, can be used as the uncertainty/confidence of the pixels, meaning that the higher the entropy value, the more informative it can be, which should receive a higher weight. Of course, there are also other means of defining the weights, such as the use of an adaptive filtering kernel [90] or through guided filtering [96], [97], as well as filters based on derived statistical hypothesis from the images themselves, such as maximum likelihood, maximum posterior, Dempster-Shafer theory, and Bayesian fusion [2], [54], [98], [100]. Many studies have suggested using image segments to guide the fusion [38], [95], [160], where the window size would change adaptively based on semantic segmentation.

### Prediction-based Methods

Pixel-level prediction considers image fusion as the pixel-level prediction problem, meaning that given a set of input images, the image fusion problem aims to predict pixel values for a new grid as the fused results. Different from the statistical filter method, the pixel-level prediction method allows to predict pixel values of a grid with a size different from the original grid, which extrapolates pixel values at cells where original information is unavailable or at a coarser granularity. Examples of pixel-level prediction methods include super-resolution [73] and spatiotemporal prediction for time-series analysis [72], [105], [108]. For instance, for super-resolution, images are up-sampled to larger sizes, creating a new mesh or gridded format of an image with old values and empty cells, which are to be filled by interpolated values calculated from neighboring pixels using methods such as spline, bicubic, Kring, etc. [46], [68], [73], [177], [178]. Typical interpolation methods operate in local domains, which are fast yet can be easily polluted by data impurities and noise of neighboring pixels. Methods like regression-based approaches aim to utilize a wider data receptive field, which can provide better inference by modeling the relationship between the raw pixel values and the predicted values [78], [179], [110], [180], [109], [74]. Kernel regression serves as one of the key approaches, where the model parameters (e.g., weights and biases) of the regression kernel can be estimated with robust estimation approaches such as robust least squares [111] with sometimes outlier removal algorithms such as random sample consensus (RANSAC) [181]. Postprocessing methods, such as Bayesian regularization [102] and Markov-Random Field [182], can be used to create fusion results under certain priors.

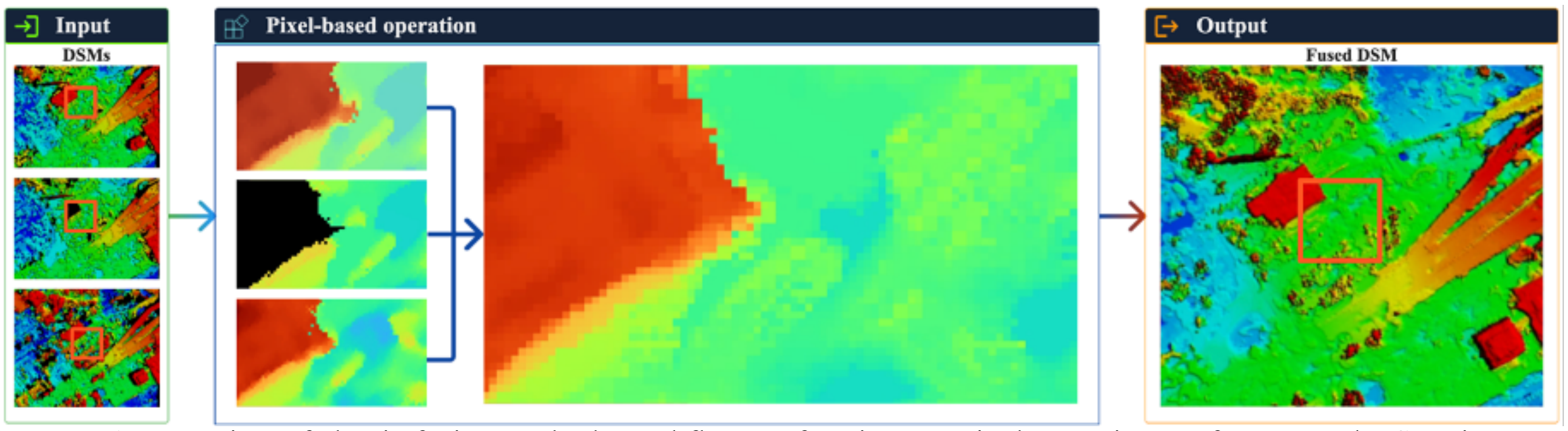

**Figure 5.** An overview of classic fusion methods workflow performing per-pixel operations to fuse several DSMs into a single DSM (as a case of M2O image fusion), where the value of the central pixel in the orange box is updated using their neighboring pixels in space and time.

**Transformation-based fusion**

Domain-transformation methods decompose images into different domains or representations and perform the data fusion under the transformed domain, such as different color space, frequency, or wavelet domains [79], [79], [183], [184]. These approaches can improve the robustness of image fusion for some tasks. For example, they can be used for spatial resolution enhancement due to their ability to preserve spatial information and fine details in images (e.g., in the frequency domain); at the same time, they are not impacted by spectral distortions. Domain transformation methods can be operated in the spatial, frequency, or other transformed domains, such as wavelets transformation [4], [76], [88], [135], [136], [139], [185]. The spatial domain includes techniques such as color transformation (e.g., Intensity-Hue-Saturation (IHS) and Brovey transformation) or projection and substitution (e.g., Principal Component Substitution (PCS), Gram-Schmidt (GS)) [4], [76], [185]. The frequency domain utilizes techniques related to the Fourier transform and its variants (e.g., Discrete Fourier Transform (DFT) or Cosine transform) [127], [132], whereas wavelet transform includes methods like continuous wavelet transform (CWT), discrete wavelet transform (DWT), as well as their variants [88], [135], [136], [139], [186]. Multi-resolution analysis (MRA) is another domain transform approach; it is a part of compressive sensing and includes methods such as curvelet transform and Laplacian pyramid transform [4], [76], [185]. These methods follow a similar processing pipeline, including image transformation, coefficient selection, coefficient fusion, and inverse transformation to the image's original domain. For example, pan-sharpening requires converting both the panchromatic and multispectral images to their frequency components using Fourier, wavelet, curvelet, or any multiscale transform-based approach. These coefficients will then be selected and fused prior to an inverse transformation. Finally, the image will be reconstructed to its original form and domain by inversing the transformation. Domain transformation-based fusion has several advantages:1) identifies surface characteristics such as roughness and repeated patterns or structures [4], [135], [186]; 2) MRA can apply fusion at multiple scales and resolutions and, thus, can effectively extract spatial details, spectral, and textural information; 3) operating in the frequency domain allows more computationally efficient analysis. Nevertheless, these methods can also introduce new artifacts as they are sensitive to noise in the input images, i.e., augmented noises in the frequency domain [135]. They also require domain expert knowledge to determine the suitable approach for a specific RS task, which may not be adaptive to all scene variations and applications.

In sum, most classic fusion methods utilize local operations, which are less effective in utilizing long-range dependencies to handle complex and non-linear inconsistencies of the inputs. In contrast, deep learning-based approaches, due to their ability to learn from examples with complex prediction models, can better address fusing tasks in removing inconsistencies caused by various and very often unknown factors (i.e., uncharacterized errors from derived products or sensor errors). Most classic fusion methods may only address fusion problems whose inconsistencies are due to assumed/relatively simple noise distributions (e.g., Gaussian distribution), whereas they may fall short when processing input images that come with inconsistency due to complex/unknown noise distributions.

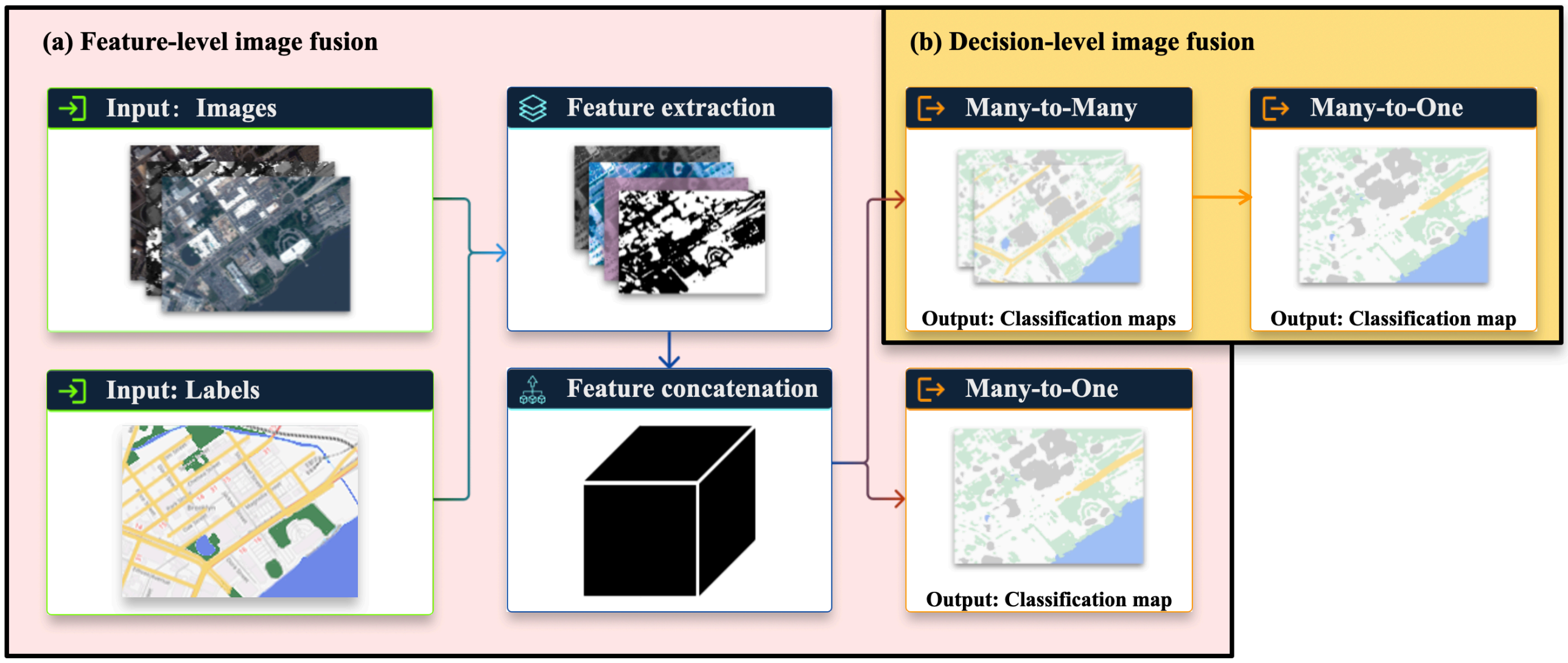

**Figure 6.** Examples of the learning-based approaches for classification that apply (a) feature-level and (b) decision-level image fusion to generate single or several classification maps as classic cases of M2M and M2O image fusion, respectively.

2.3.2. Learning-based Fusion Methods

Learning how to produce fused results through examples via means of ML and DL has become increasingly attractive for many image fusion tasks. The conventional ML fusion methods utilize algorithms through simpler models such as support vector machine (SVM), random forest, decision trees, etc. [36], [58], [187], which learn either parameters for statistical filtering, or pixel-level prediction by examples in applications such as super-resolution and time-series analysis. In comparison, DL methods apply fusion through more complex models such as convolutional neural networks (CNNs), generative neural networks, recurrent neural networks (RNNs), attention mechanisms, and transformers [169], [188], [86], [87], [153], [158], [189]. Learning-based methods are data-driven, meaning they can extract features and learn patterns directly and automatically from images, enabling image fusion to be performed mostly on the feature and decision levels. The feature-level fusion is where learned features (i.e., feature vectors or feature maps) are extracted separately from images and fused in the following steps, where decision-level fusion combines the predicted images as a postprocessing step (See Figure 6). As is the common problem in ML/DL, learning-based approaches have limited transferability and generalization to new scenes or datasets due to large domain gaps between training and testing images [190], [191], [192], [193], [194]. Oftentimes, there are also limited RS-based training data, which can impact the performance of such models.

**Conventional machine learning (ML) methods for image fusion**

ML methods are prevalent for image fusion tasks [148], [149], [195], [196], which are primarily divided into feature-level and decision-level fusion. The feature-level fusion performs linear or statistical fusion at the feature space (i.e. hand-crafted feature) in the prediction pipeline [13], [14], [15], [37], [65], [197], the key of which is to learn feature descriptions of multi-modal data into a space where these features can be easily fused without creating conflict of evidence [3], [4], [149], [195]. Feature-level fusion is mostly used in multisource and multimodal data-based fusion, which serves as a means to effectively utilize complementary information for typical tasks such as semantic segmentation, instance segmentation, etc. The decision-level fusion, as what it names, performs fusion on the predicted results, such as from classification/semantic segmentation maps. As compared to feature-level fusion, decision-level fusion is made simpler because the predicted results are already in the same modality with simple fusing strategies such as weight fusion or majority voting, averaging, median, etc. [44], [58], [60], [62], [198]. Of course, the downside of the decision-level fusion is the lack of flexibility to infer earlier processes in feature selection. In sum, both approaches dominated classic fusion literature when handling multi-modal data.

**Deep learning (DL) methods for image fusion**

DL models possess significant advances in learning complex features and pixel-level prediction, as shown in recent literature [84]. While technically, it introduces methodological complexity, it mostly differs from conventional ML by using models that are highly parameterized, whereas the concept of fusion strategies remains, such as the data-level, feature-level, or decision-level fusion. However, unlike ML methods, DL methods can automatically extract low to high-level features without manual interference; however, they may require significant training data and time for learning and processing images. From an application perspective, DL methods are mostly used in two types of applications: spatial-spectral image fusion that focuses on M2O and spatiotemporal fusion that produces predictions through time (M2M). The following subsections entail relevant works in these two types.

a. Deep learning for spatial-spectral image fusion

As mentioned above, DL serves as a more complex mapper in image fusion. One key application in RS image fusion is to create a spatially and spectrally enhanced image from a set of spatially/spectrally low-resolution images (M2O fusion). Earlier success in DL fusion use CNNs and their variants, such as ResNet, U-Net, and AlexNet [47], [153], [154], [158], [199], which can efficiently extract not only spatial details (e.g., 2D shapes), but also valuable contextual information. For example, Pan-sharpen fuses a low-resolution MS image with a high-resolution PAN image into a high-resolution MS image. Typically, the conventional approach explicitly extracts and augments high-frequency details to reserve in the final image while blending spectral information in the low-frequency domain. This process will consequently augment noises from the original image. DL models, on the other hand, can segregate noise from spatial details by learning from examples, such that when fusing the MS and PAN images into the Pan-Sharpened image, unwanted noises are suppressed to minimum. Due to the higher model capacity, DL models can also directly learn, again from examples, predicting pixel-level results directly from MS and PAN images, which learns complex feature representations that best fit the prediction results towards the ground-truth. For instance, in the work of [155], they performed super-resolution on an MS image using super-resolution CNN (SRCNN), then combined the new high-resolution MS with the PAN image using the GS method (a classic fusion method) to get more accurate spatial and spectral representation. A similar line of work can be found in the application of super-resolution [156], [200], [201] and multi-modality fusion based learning as well [202], [203].

b. Deep learning for spatiotemporal image fusion

DL also benefits the spatiotemporal image fusion by providing powerful connections for modeling long-range dependencies in both the spatial and the temporal direction, advancing applications such as time-series analysis, change detection, and crop and vegetation dynamics prediction [147], [156], [163]. Well-parameterized models such as recurrent neural networks (RNNs), Long Short-term Memory (LSTM), and Transformers have been widely used in such applications. These types of networks have a built-in structure to store and reuse information with an increasing range of dependencies (from RNN to transformers) [86], [189], [204], [205], [206], [207], and thus they are capable of performing spatiotemporal image fusion considering well-modeled trend for improving the fusion accuracy. For example, typical spatiotemporal filters only consider local neighborhood tensors and are parameterized by simple one-pass linear weights and biases, which lack the capability to model complex phenological patterns. In contrast, DL models with millions of parameters can easily do [84]. The three typical DL models accounting for temporal dependencies, i.e., RNN, LSTM, and Transformer, differ in their structure, mechanism, and memory length due to their design. Thus, fusion can be applied at different stages depending on the network. For instance, RNNs operate in a loop structure, where important information is stored from the first input image (in what is called a hidden state) and fused with the following image to be processed in an RNN cell, and so forth until the network is set to terminate at a given condition or when all images are consumed. This gives the opportunity to fusion tasks to retain pixel information that is static throughout the time. LSTM and Gated Recurrent Unit (GRU) have similar procedures; however, they have a longer memory to store temporal dependencies and contextual information and use what are called gates to control input, output, and information flow [199]. However, the range of dependencies in the sequential data (e.g., multi-temporal images) is still limited. Recently, researchers explored alternatives that can leverage performance and resources, especially for geographical data, which can be in the form of self-supervision. This led to the adoption of attention mechanisms through transformers, a seminal work that outperforms traditional DL networks in incorporating global dependencies, and thus, it can be naturally used in image fusion tasks [87], [169], [170], [171], [172], [173], [174], [205], [208], [209], [210]. Transformer, in adaptation of the attention mechanism through even longer-range dependencies, shows its ability to exploit global information, which outperforms RNN/LSTM-based structures. Moreover, transformers can also extract features in both the spatial and temporal directions. For example, ViT (Vision Transformer) divides images into multiple smaller patches to allow the attention mechanism (cross-attention) in Transformer to explore global information within the image [170], [171], [172], [173], [174], [205]. Similarly, cross-attention can be applied to divided patches among images captured at different times to allow for temporal information encoding [171]. In sum, the use of transformers for fusion is still growing, and more such methods are expected to be developed soon.

## 3. A Meta-analysis of Existing Works Using Remote Sensing Fusion as a Core Technique

We aim to review the topic of remote sensing image fusion in a broad sense; thus, we have included a meta-analysis on papers that involve image fusion as a core technique, aiming to analyze the impact of image fusion on the field of RS. The meta-analysis is designed to help inform existing and most recent practices of image fusion methods in the RS field. On the one hand, this helped us to develop our new taxonomy that divides image fusion based on the strategy and the "input-output" formation; on the other hand, it helped us understand the most recent fusion methods that have not been investigated yet in previous review literature. Therefore, for our meta-analysis, we will first present our search method, while in the next Section (Section 4), we will present the results of the statistical analysis showing the development of fusion methods and applications over the years.

### 3.1. Data Selection and Extraction

In this review, we searched for articles under the remote sensing image fusion topic using the Scopus database (as of 2023), where we set a few queries and searched them in the title, abstract, and keyword sections. The search query includes a few sentences like "Remote sensing" AND "Image fusion", "Remote sensing" AND "Fusion", "Fusion" AND "Digital Surface Models" AND "Fusion", and "Digital Elevation Models" AND "Fusion" (See Table II). We then refine the search by picking only peer-reviewed journal and conference articles and voiding everything else. The initial results led to a total number of 5,926 articles, which we further sorted into journal or conference articles that count for 3,466 and 2,460 articles, respectively.

**Table II. The queries of this review**

| Query |
|---|
| "Image fusion" AND "Remote sensing" |
| "Remote sensing" AND "Fusion" |
| "Digital Surface Models (DSM)" AND "Fusion" |
| "Digital Elevation Models (DEM)" AND "Fusion" |

3.2.Data Information Sources

Our search only includes journal and conference articles that have been peer-reviewed and are well-recognized in the remote sensing society. We summarize these articles in the list shown in Table III. Our list comes from several sources, such as the IEEE Geoscience and Remote Sensing Society (IGARSS), the annals of the International Society for Photogrammetry and Remote Sensing (ISPRS), the Society of Photo-Optical Instrumentation Engineers (SPIE), etc. Table III indicates that the total number of published conference articles is about 2,460, whereas the list of peer-reviewed articles from well-established journals is about 3,466.

**Table III. The titles of the conferences and journals selected for the review with the number of published articles concerning fusion in remote sensing (as of 2023)**

| Conference Name | # |
|---|---|
| International Geoscience and Remote Sensing Symposium IGARSS | 1,225 |
| Proceedings of SPIE The International Society for Optical Engineering | 935 |
| IEEE ISPRS Joint Workshop on Remote Sensing and Data Fusion Over Urban Areas DFUA 2001 | 68 |
| 2nd GRSS ISPRS Joint Workshop on Remote Sensing and Data Fusion Over Urban Areas Urban 2003 | 64 |
| ISPRS Annals of The Photogrammetry Remote Sensing and Spatial Information Sciences | 57 |
| Proceedings International Conference on Image Processing ICIP | 27 |
| 2011 International Conference on Remote Sensing Environment and Transportation Engineering RSETE 2011 Proceedings | 18 |
| Proceedings Applied Imagery Pattern Recognition Workshop | 12 |
| IEEE Aerospace Conference Proceedings | 12 |
| International Conference on Signal Processing Proceedings ICSP | 11 |
| IEEE Computer Society Conference on Computer Vision and Pattern Recognition Workshops | 9 |
| Digest International Geoscience and Remote Sensing Symposium IGARSS | 8 |
| Conference Record IEEE Instrumentation and Measurement Technology Conference | 7 |
| IEEE Conference on Intelligent Transportation Systems Proceedings ITSC | 7 |
| **Total number of conference publications** | **2,460** |
| **Journal Name** | **#** |
| Remote Sensing | 837 |
| IEEE Transactions in Geoscience and Remote Sensing | 619 |
| IEEE Journal of Selected Topics in Applied Earth Observations and Remote Sensing | 419 |
| IEEE Geoscience and Remote Sensing Letters | 320 |
| International Journal of Remote Sensing | 288 |
| Remote Sensing of Environment | 166 |
| ISPRS Journal of Photogrammetry and Remote Sensing | 142 |
| Journal of Applied Remote Sensing | 121 |
| International Journal of Applied Earth Observation and Geoinformation | 99 |
| Photogrammetric Engineering and Remote Sensing | 62 |
| ISPRS International Journal of Geo-Information | 54 |
| Information Fusion | 46 |
| Applied Sciences Switzerland | 45 |
| International Journal of Image and Data Fusion | 34 |
| Journal of Geo-Information Science | 31 |
| Remote Sensing Letters | 31 |
| Canadian Journal of Remote Sensing | 26 |
| IEEE Transactions on Image Processing | 22 |
| European Journal of Remote Sensing | 18 |
| IEEE Transactions on Instrumentation and Measurement | 17 |
| IEEE Transactions on Neural Networks and Learning Systems | 17 |
| Computers and Geosciences | 13 |
| IEEE Sensors Journal | 11 |
| Journal of Environmental Management | 10 |
| Photogrammetric Record | 9 |
| Pattern Recognition Letters | 9 |
| **Total number of journal publications** | **3,466** |
| **Total number of conference and journal publications** | **5,926** |

## 4. META-ANALYSIS RESULTS

This section presents the meta-analysis results for the selected data from the previous query, as described in Section 3. We will show the word cloud for the keywords from our search, the frequency of annual journal and conference publications, the number of publications per country, and the frequency of publications discussing specific RS applications.

4.1. Statistical Analysis for the Keywords in this Review

Figure 7 shows the main keywords as they appear in the selected articles. Words with large fonts indicate their appearance in high frequency and vice versa for words with smaller fonts. Our results in Figure 7 suggest that the words "Image fusion" and "Data fusion" are the most frequent words to appear. The figure also indicates that the most applications using image fusion are classification, spatial resolution enhancement, and image enhancement, which are applied to images such as satellite images such as MODIS, hyperspectral images, and classification maps.

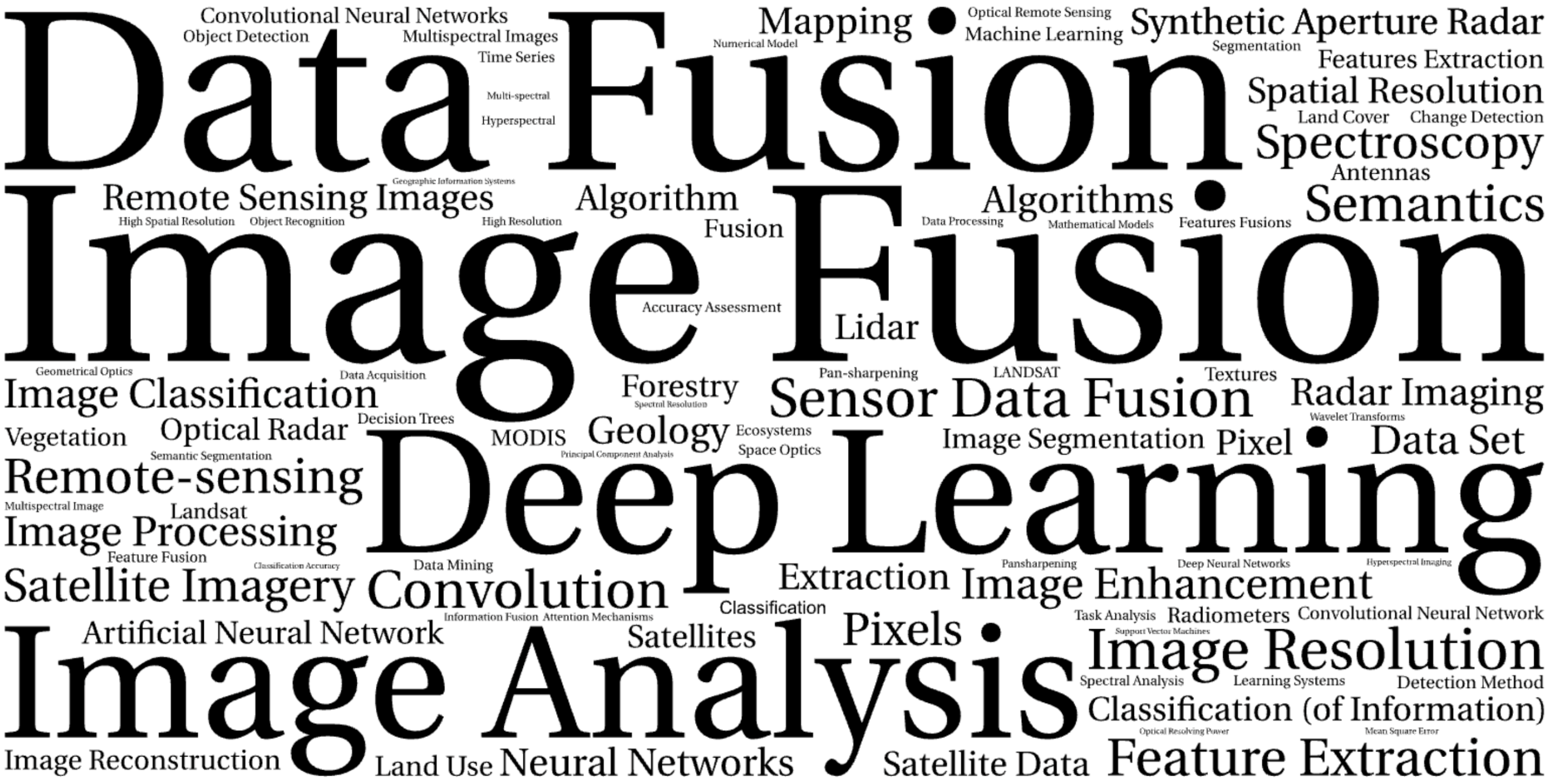

**Figure 7.** A visualization of the word cloud for the keywords of this review

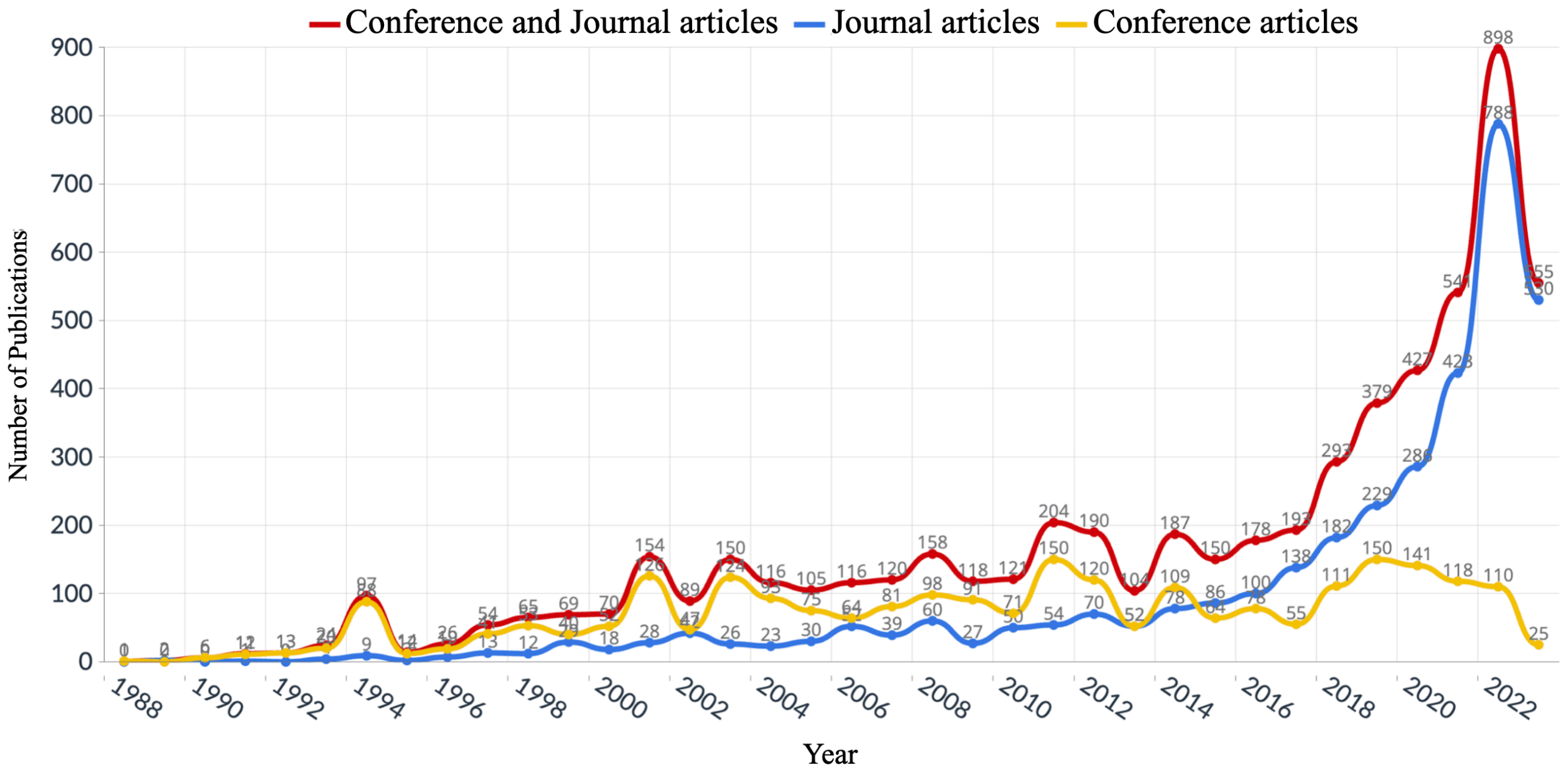

**Figure 8.** The statistical analysis of the annual conference and journal publications.

4.2. Statistical Analysis of the Annual Publications

In Figure 8, we can see the pattern of the number of articles published over the years. We compare the total number of articles from journals, conferences, and both, as demonstrated by the red, blue, and yellow lines. The figure shows that image fusion has been in the field for decades since the 1980s. It continues to grow gradually over the years from several hundred publications to almost a thousand publications per year, indicating that more research was proposed to enhance the fusion performance in RS (see the red line in Figure 8). The number of conference articles was dominant up to 2015, as seen from the yellow line in Figure 8. After that, the number of journal articles significantly increased over the years, as indicated by the blue line in Figure 8.

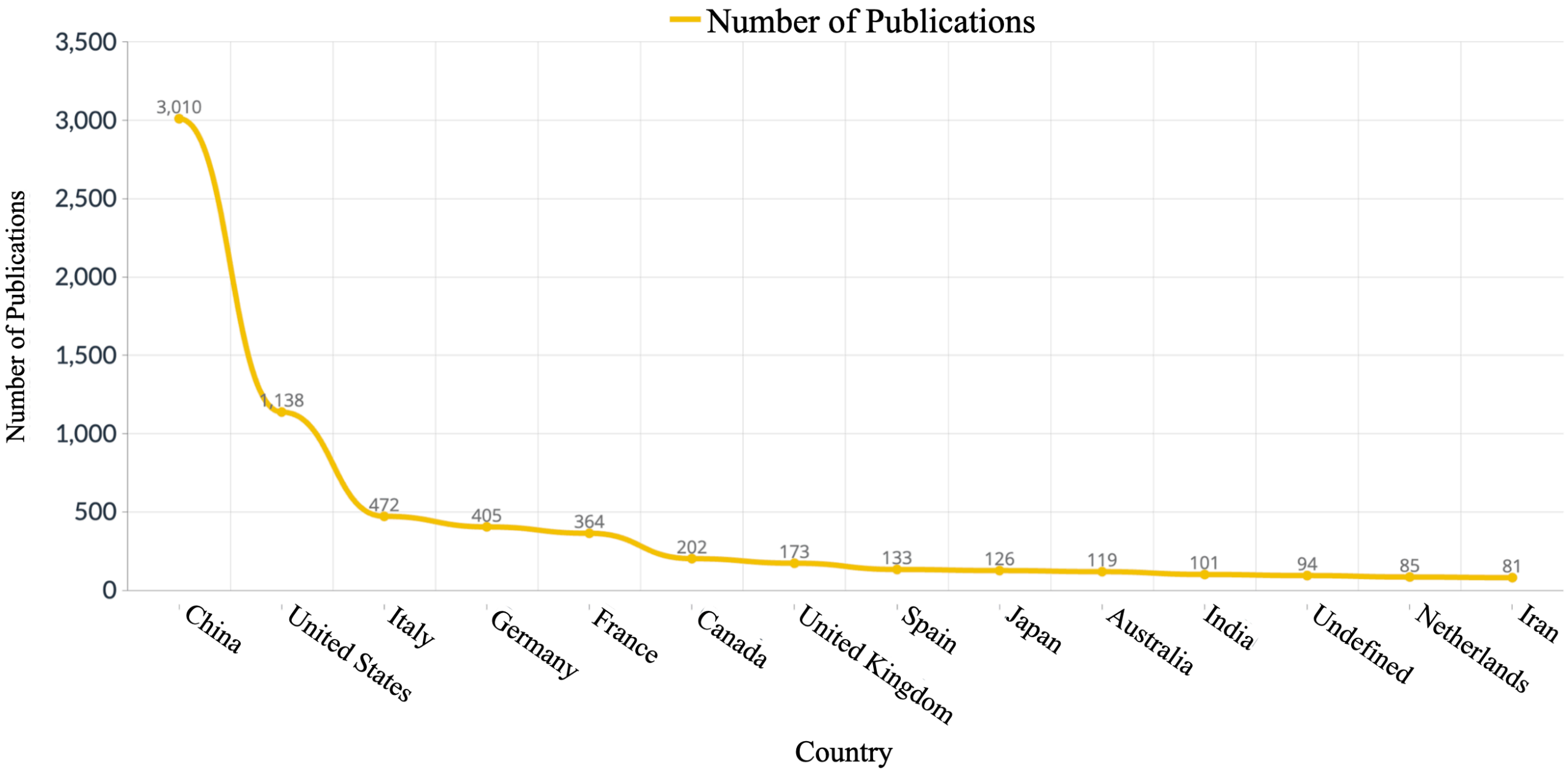

**Figure 9.** The statistical analysis of the number of publications per country. Note: the "undefined" in the x-axis refers to unknown origins or countries of the authors.

### 4.3. Statistical Analysis of the Publications Based on Country

Researchers from 103 countries worldwide have published at least one journal or conference paper related to the image fusion topic. In Figure 9, we provide a plot that summarizes the top 14 countries with the highest number of publications on this subject. The figure shows that China makes at least 50% of the contribution towards this field compared to all other countries, with the number of publications of around ~3,000. The United States of America (USA) also has a high record of publications concerning image fusion in RS, with around ~1,100 articles. Italy, France, Germany, Canada, United Kingdom, Spain, Japan, Australia, and India have a lower publication rate ranging between 100 and 500. The remaining countries still contribute to this field but with a lower number of publications, less than 100.

### 4.4. Statistical Analysis of the Approaches and Applications of Image Fusion

Image fusion methods, as explored in Section 2.3, are broadly categorized into classic and learning-based methods. Classic methods encompass statistical, numerical, and domain transform techniques [58], [77], [69], [78], [79], [80], [43], while learning-based methods incorporate ML and DL approaches [65], [66], [36], [47], [62], [85], [86], [87]. The progression of the number of publications concerning the development of the classic (yellow line) and learning-based (blue line) image fusion methods over the past 30+ years is shown in Figure 10, which highlights a significant difference. Learning-based methods have shown a remarkable increase, reaching nearly ~1,000 publications in 2022 alone, in contrast to the classic methods, which have maintained a steady output of no more than 200 publications annually. DL methods have demonstrated a dramatic rise in contributions since their establishment in 2015. Two additional lines in Figure 10 illustrate this: the purple line for ML and the green line for DL methods. While both ML and DL methods initially had similar growth in publications, after the year 2015, DL (purple line) publications significantly overtook those of ML (green line) and, by the year 2020, surpassed classic methods as well. This underscores the breakthrough of DL in image fusion post-2015, though classic and ML methods continue to be integral in many applications.

In terms of applications, image fusion is pivotal in Remote Sensing (RS) with diverse uses categorized into six main tasks: 1) Classification refinement, including supervised classification, land use land cover (LULC), and segmentation, and any other application related to refining the accuracy and precision of the classification maps [58], [65], [60], [66], [36], [12], [62], [40], [70], [198], [63]; 2) Detection and recognition, such as building and change detection [64], [179], [56], [45], [211], [212], [123]; 3) Spectral image enhancement for spectral coherence, denoising, filtering, and restoration [25], [213], [116], [22], [214], [69], [215], [23], [72]; 4) Spatial resolution enhancement for tasks like pan-sharpening or super-resolution [6], [47], [86], [216], [68], [217]; 5) 3D reconstruction involving DSM and DEM fusion, and stereo matching [43], [49], [75], [85], [152], [218]; 6) Environmental studies for monitoring and managing natural events or disasters [66], [169], [219], [220], [221], [222]. These applications are summarized in Figure 10. The majority of research focuses on refining classification maps (nearly 30% of total publications), followed by environmental studies (approximately 26.2%), which includes environmental monitoring and disaster management. Spatial resolution enhancement, 3D reconstruction, and spectral image enhancement have comparable publication shares (ranging from 11-14.5%), while detection and recognition hold about 7% of total publications.

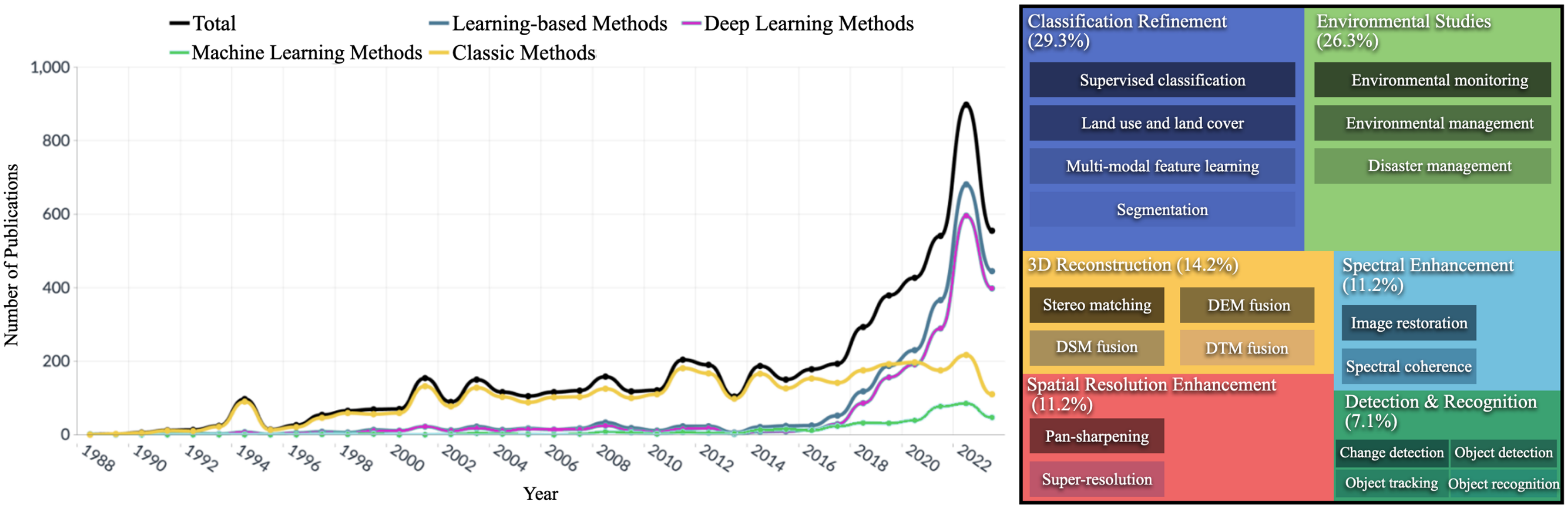

**Figure 10.** An overview of image fusion approaches and applications covered in RS shows the publication numbers and percentages over the years. The total number of articles surveyed is 5,926.

## 5. Summary and Future Directions

### 5.1. Summary and Conclusion

In this article, we have reviewed the topic of image fusion for almost all areas in the RS field, i.e., images of different types, fusion techniques, and applications. We focused on providing a broader overview of the fusion problem, where we have introduced a new taxonomy that manifests the two most common categories of image fusion as practiced in the RS field, which includes M2O and M2M image fusion. This taxonomy, although it is simple in notion, is fundamental as it presents fusion as a mapping function that can process different numbers of inputs and outputs. Therefore, the fusion strategies can be extended to augment different applications based on similar concepts and not be limited to specific types of images (i.e., spectral) or applications (i.e., pan-sharpening) as presented in previous literature. Moreover, we have presented a meta-analysis that covered 5,926 peer-reviewed articles to provide a statistical analysis showing the progression of image fusion methods and applications over the years. The meta-analysis indicates a significant growth in image fusion methods, especially with the introduction of DL methods, which has boosted the number of publications from several hundred to almost 1,000 in a year. The statistical results also indicate that classification, environmental studies, and 3D reconstruction are the top three applications that benefit from image fusion, where they account for almost ~15-30% of all publications. On the other hand, applications related to spectral and spatial resolution enhancement account for almost ~11% of all publications. This conclusion leads to rethinking the past literature reviewing RS image fusion, which almost described the fusion problem as an approach for spatial resolution enhancement, specifically for pan-sharpening. Our meta-analysis, on the other hand, indicates that fusion methods can be applied to many types of images and applications. Therefore, we explicitly describe and present the main categories of image fusion methods and their RS applications to show the unlimited capabilities of such strategies.

### 5.2. Discussion and Future Directions

Image fusion has been a long-investigated topic, yet it has grown to become increasingly complex due to the ever-developing sensory data and diversified inputs (sources, derived products, different modalities). Therefore, existing syntheses focused on their sub-topics due to the divergent applications associated with image fusion, such as image super-resolution, spatiotemporal filtering, and pan-sharpening. In our investigation, we re-synergize these works back to the fundamentals of fusion by categorizing the image fusion through two simple strategies, i.e., many-to-one (M2O) and many-to-many (M2M) image fusion. With an overview, we describe typical strategies and used models related to these general tasks to provide a starting point overview for researchers interested in image fusion approaches and applications, thus fueling future research. Obviously, due to the highly disparate image-fusion tasks, specific considerations should be accounted for when addressing specific image-fusion challenges. At the same time, with the technological developments (both in sensors and algorithms), there is a converging trend of using deeper models for image-fusion tasks. Specific conclusions and recommendations are as follows:

**Knowing the cause of inconsistency:** When preparing images for fusion, i.e., many-to-one (M2O) or many-to-many (M2M) applications, it is essential to understand the tasks and causes of inconsistency for devising the fusion algorithms. Section 2.1 entails various examples of the causes of inconsistency; thus, specific algorithms addressing such challenges in fusion applications should be considered. Moreover, M2O algorithms emphasize information homogeneity among different inputs, while M2M algorithms focus on information compensation and signal-noise separation due to their task of using multiple images to augment each other. These constitute the core challenges to address, which are being tackled by both classic and various deep learning models (Section 2.2-2.3)

**Multi-modality learning:** The recent advance of learning models suggests a trend towards using multi-modality remote sensing data for fusion, specifically, the use of Radar, optical, and cross-modal input such as GIS (Geographical Information System) data. It is seen that the fundamentals of these tasks are the fusion algorithms in yielding informative remote sensing assets for better interpretation, such as for information retrieval via augmented spatial-temporal resolution, as well as unique 2D/3D signatures (such as thermal, cloud-penetrating Radar, as well as LiDAR) for improved classification, detection and quantification of ground objects. Many of these tasks serve critical environmental and civilian applications.

**Trends and future directions**: Our meta-analysis over 5,926 articles (Section 3-4) yields several conclusions: first, the use of image fusion appears to be highest in publications related to the sub-domain of "remote sensing and geosciences", where the fusion of images for image interpretation is critical. Second, image fusion is mostly used in applications for image classification and environmental studies, more than 50% of the published works. The rest are shared (roughly equal) among 3D fusion, spatial/spectral enhancement, and detection. Third, there is a converging trend of using deep models for image fusion, while classic methods, although taking a significantly smaller fraction of published works, remain a fair number as compared to previous years, noting that there is still a good use of older methods, as they require much fewer labels. Although with a fairly smaller number, we note that there is increasing work on building foundational models that could interact with various remote sensing tasks, including image fusion and the fusion of cross-modality data, such as the use of images, texts, and audio via using large language models [223], [224], [225].

In sum, our review suggests that image fusion, or the fusion of gridded data, still plays a vital role in this era, where both sensors and models are developed at a rapid pace. In addition to improving classic image fusion algorithms to accommodate newer sensory data and derived data (e.g., DSMs), there has been a paradigm shift from canonical methods to using complex learning and generative models to yield more accurate and comprehensively fused results. Although there has been less focus on already-used algorithms such as pan-sharpening, there is a large room for development when involving multi-modality, non-raster data.

## Acknowledgment

Rongjun Qin is partially funded by the Office of Naval Research (grant numbers N000142012141 & N000142312670).